%% file: acl_latex.tex
\documentclass[11pt]{article}

\usepackage[final]{acl}

\usepackage{times}
\usepackage{latexsym}

\usepackage[T1]{fontenc}

\usepackage[utf8]{inputenc}
\usepackage{microtype}
\usepackage{marvosym}
\usepackage{inconsolata}

\usepackage{graphicx}

\usepackage{amsmath}
\usepackage{amssymb}
\usepackage{amsthm}

\usepackage{algorithm}
\usepackage{algorithmicx}
\usepackage{algpseudocode}  

\usepackage{booktabs}  
\usepackage{multirow}  
\usepackage{array}    

\usepackage{subfigure}

\usepackage{xcolor}
\usepackage{listings}
\usepackage{xcolor}

\usepackage[most]{tcolorbox}
\usepackage[T1]{fontenc}

\usepackage{hyperref}
\usepackage{algpseudocode}
\usepackage{url}
\usepackage{natbib}

\title{SEAD: Self-Evolving Agent for Multi-Turn Service Dialogue}

\author{
    Yuqin Dai,
    Ning Gao,
    Wei Zhang,
    Jie Wang,
    Zichen Luo,\\
    Jinpeng Wang, 
    Yujie Wang , 
    Ruiyuan Wu ,
    Chaozheng Wang\textsuperscript{\Letter}\\
    Meituan \\
   \small{\textbf{\Letter Correspondence:} \href{mailto:adf111178@gmail.com}{adf111178@gmail.com}}
}

\begin{document}
\maketitle
\begin{abstract}
Large Language Models have demonstrated remarkable capabilities in open-domain dialogues. 
However, current methods exhibit suboptimal performance in service dialogues, 
as they rely on noisy, low-quality human conversation data.
This limitation arises from data scarcity and the difficulty of simulating 
authentic, goal-oriented user behaviors.
To address these issues, we propose \textbf{SEAD} (\textbf{S}elf-\textbf{E}volving \textbf{A}gent for Service \textbf{D}ialogue), a framework that enables agents to learn effective strategies without large-scale human annotations. SEAD decouples user modeling into two components: a Profile Controller that generates diverse user states to manage training curriculum, and a User Role-play Model that focuses on realistic role-playing. This design ensures the environment provides adaptive training scenarios rather than acting as an unfair adversary. 
Experiments demonstrate that SEAD significantly outperforms Open-source Foundation Models and Closed-source Commercial Models, improving task completion rate by 17.6\% and dialogue efficiency by 11.1\%. 
Code is available at: \url{https://github.com/Da1yuqin/SEAD}.
\end{abstract}

\section{Introduction}

Large Language Models (LLMs) have revolutionized a wide range of applications across diverse domains~\cite{zhang2025ncvnodewiseconsistencyverification, zhang2026consensusentropyharnessingmultivlm, dai2025careful, zeng2025card, yin2025floorplan}. 
However, training robust goal-oriented dialogue agents remains bottlenecked by scarce, expensive, and low-quality conversational data \cite{hosseini2020simple}. In multi-turn service dialogue, agents must dynamically track user states and adapt strategies across extended interactions. 
Human logs from sales calls are a perfect example of these challenges \cite{qian2022toward}: agents lack standardization, requiring costly filtering, and the final data is heavily lopsided towards failed attempts.
Therefore, data quality is fundamentally capped by the capabilities of the original human agents who produced these conversations.

\begin{figure}[t]
    \centering
    \includegraphics[width=\linewidth]{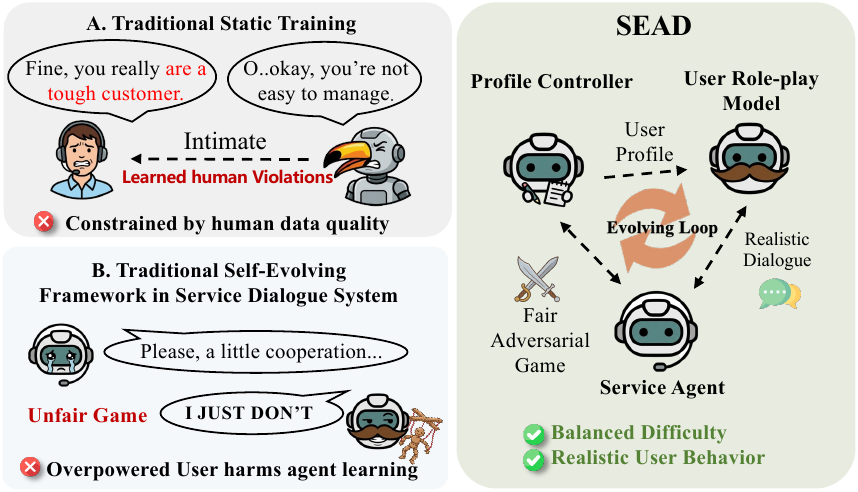}
    \caption{
    Comparison of training paradigms. A. Static data methods are limited by data quality and may learn human violations. B. Traditional self-evolving creates unfair games where user agents dominate outcomes. Our \textbf{SEAD} achieves balanced co-evolution and realistic interactions through decomposed user modeling, forming a fair adversarial game.
    }
    \label{fig:mainfig}
\end{figure}

To solve data quality issues, prior works~\cite{li2025wizard, ou2024dialogbench} have explored synthetic data and user simulation as alternatives. Static synthesis approaches \cite{ou2024dialogbench, 10.1145/3786793} generate fixed datasets from predefined scenarios but cannot capture dynamic conversations where users react to agent behavior. Interactive simulation attempts typically employ LLMs to dynamically generate user responses during agent training \cite{sekulic2024reliable, zhu2025evaluating, gao2026reinforcing, sage}, but face critical limitations: without perceiving the agent's current capability level, simulators either generate overly challenging scenarios that frustrate learning (too strong) or overly simple scenarios that provide insufficient training signal (too weak), both yielding suboptimal training outcomes. Additionally, existing simulators exhibit overly perfect styles \cite{lin2025user}, lacking real users' attention lapses, linguistic noise, and irrationality \cite{takanobu2020your}. Most critically, both fixed datasets and non-adaptive simulators fail to provide curriculum-based training that adjusts difficulty as the agent improves, limiting their effectiveness for complex multi-turn service dialogues requiring coherent long-term interactions.

Recent advances in self-evolution offer promising zero-data solutions for data-scarce business scenarios. These self-evolving methods\cite{chen2024self, zhao2025absolute} employ self-play strategies where models generate both questions and corresponding answers, eliminating dependence on expensive expert-annotated data ~\citep{silver2017mastering,vinyals2019grandmaster}. 
However, applying these methods to service dialogues faces critical challenges. As shown in Figure~\ref{fig:mainfig}, first, \textbf{unfair adversarial game}: in the service dialogue scenario, User Role-play Models can arbitrarily control outcomes, breaking the causal link between agent actions and task success. For example, simulators may reject agents regardless of response quality or accept based on turn count rather than persuasion effectiveness. Second, \textbf{real users are highly diverse}: without additional mechanisms, user behaviors easily fall into repetitive patterns.

To address these challenges, we propose SEAD (Self-Evolving Agent for Dialogue), the first self-evolving framework for multi-turn service dialogues. SEAD requires no large-scale annotated dialogue data, only user profiles and standard operating procedures as inputs.
To avoid the unfair adversarial game where User Role-play Models arbitrarily control outcomes, we decompose the user side into two components: a profile generator that samples initial user states, and a role-play model that simulates responses. Crucially, only the profile generator participates in adversarial training by setting initial conditions. 
This design transforms participation into a betting game, where the user side must genuinely consider agent's capability to identify the \textit{golden training scenarios} where agents can succeed approximately half the time, enabling genuine adversarial learning.
To maintain user diversity, the profile generator employs automated random sampling and consistency checks to ensure scenario diversity and authenticity.

We validate SEAD in outbound call services based on a real enterprise scenario. By enumerating 5 cooperation levels, 4 emotion levels, and 6 trust levels, we construct 120 initial user state combinations, evaluated through multi-level metrics. Experiments show SEAD significantly outperforms baselines using foundation models or large APIs. Small models trained via SEAD achieve superior performance while drastically reducing costs. Notably, SEAD remains effective in data-scarce domains, enabling rapid deployment.

Our main contributions include:
\begin{itemize}
    \item We propose the first self-evolving framework for multi-turn service dialogues that requires no large-scale annotated dialogue data.
    \item We design a decomposed user modeling mechanism that transforms participation into a betting game, forcing the user side to identify golden training scenarios and enabling genuine adversarial learning.
    \item We design a user scenario generation mechanism based on anonymized behavior patterns extracted from over 100k real dialogues, ensuring diversity, authenticity, and adaptive difficulty.
    \item Experiments show that SEAD achieves superior performance with significantly smaller model size: better at guiding users toward goals, more efficient in conversation flow, stronger in understanding user states, and more realistic in simulating authentic user behaviors, all without requiring large-scale annotated dialogue data.
\end{itemize}

\input{sections/RelatedWorks}

\section{Methodology}\label{sec:methodology}
\begin{figure*}[t]
    \centering
    \includegraphics[width=0.97\textwidth]{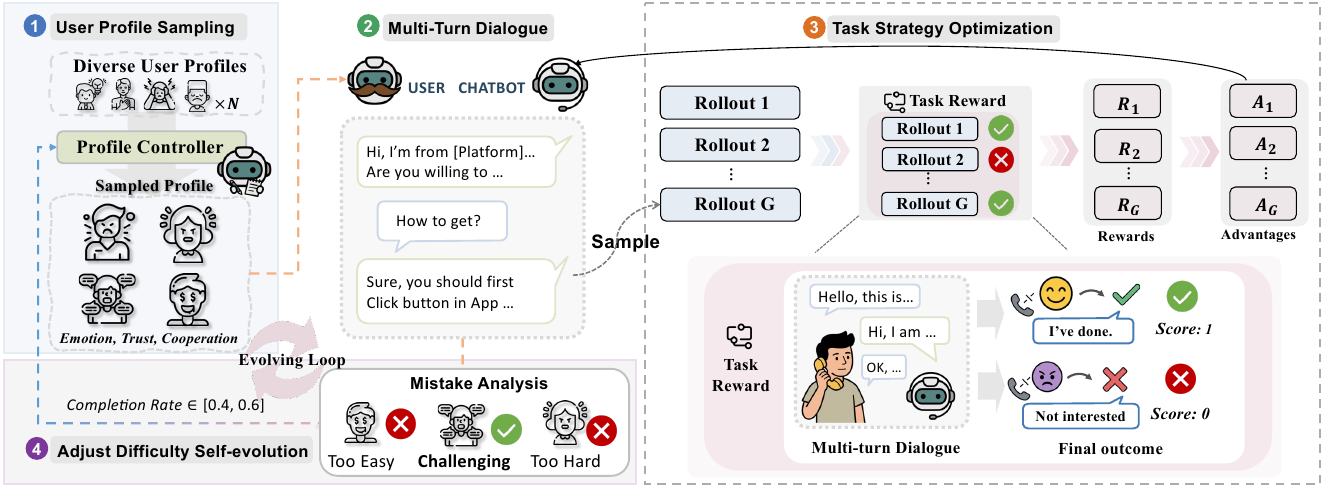}
    \caption{SEAD Co-evolutionary Training Loop. The controller samples initial states (Phase 1), which initialize dialogues producing trajectories (Phase 2), used to train the agent with rewards (Phase 3) and compute completion rates (Phase 4), which feed back to adjust sampling distributions, closing the co-evolutionary loop.}
    \label{fig:framework}
\end{figure*}

Service dialogue faces severe data scarcity, making self-evolving frameworks a promising solution. However, unlike objective tasks where correctness is verifiable, service dialogue outcomes are entirely subjective—users can arbitrarily control results regardless of agent quality, creating an unfair adversarial game. To resolve this, SEAD decouples user modeling into two components: a Profile Controller that samples initial states and participates in adversarial training, and a User Role-Play Model that focuses on realistic simulation without controlling outcomes. This design guides the Profile Controller to identify golden training scenarios (agent success rate $\sim$50\%) through initial state selection rather than mid-dialogue manipulation. Figure~\ref{fig:framework} illustrates the complete framework with four training phases. We first formalize the problem and define notations, then detail the framework components and training process.

\subsection{Problem Formulation and Framework Components}

\begin{figure}[t]
    \centering
    \includegraphics[width=0.95\linewidth]{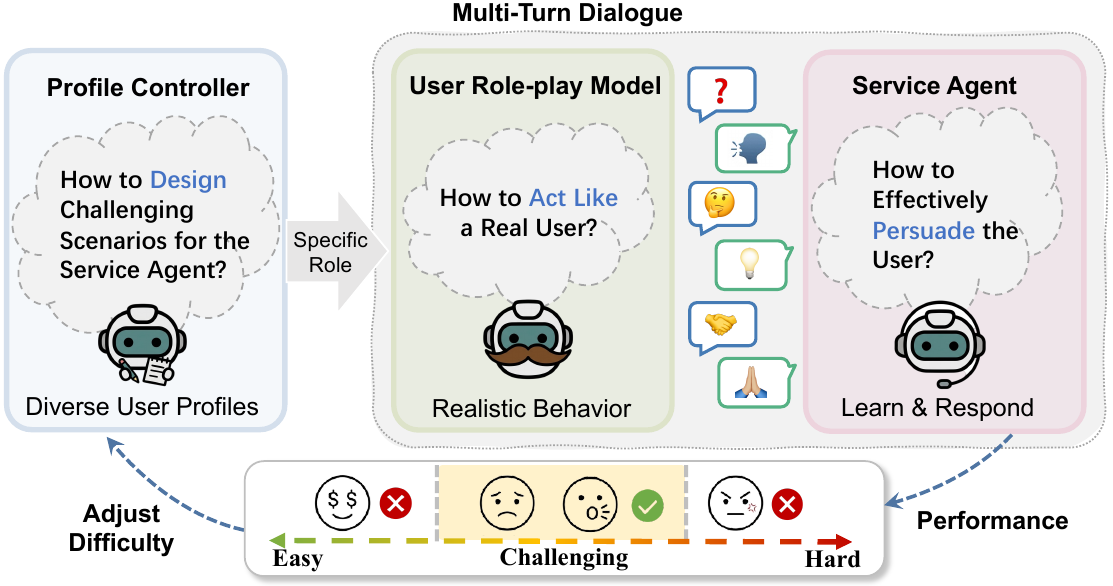}
    \caption{SEAD Framework Overview. SEAD consists of three components: (1) \textbf{Profile Generator} first creates diverse user profiles, then the (2) \textbf{User Role-Play Model} enacts these users to interact with the (3) \textbf{Service Agent}, training agents to adapt to any user. 
    Finally, these dialogue data reflecting service agent capability returns to the Profile Controller, and initiates the next evolving loop.
    }
    \label{fig:components}
\end{figure}

We aim to train a service agent that maximizes the reward $R$, measuring task completion and user satisfaction. Figure~\ref{fig:components} illustrates our framework.
First, the profile generator $\pi_g$ samples initial user states $p_0$ to create diverse user profiles, where $p_\theta$ denotes the parameterized state distribution. Then, the user role-play model $\pi_u$ enacts this user to interact with the service agent $\pi_a$ through multi-turn dialogues.
We model multi-turn service dialogue as a sequential decision process. At each turn $t$, the agent observes dialogue history $h_t = \{u_1, a_1, \ldots, u_t\}$ and generates response $a_t$, where user response $u_t$ and agent action $a_t$ alternate. The agent maintains state estimate $\hat{s}_t$ to guide action selection, maximizing cumulative reward $R = \sum_{t=1}^{T} r_t$. User state $p_t = (c_t, e_t, tr_t)$ represents cooperation $c_t$, emotion $e_t$, and trust $tr_t$, which evolve based on agent behavior. User profile $p_0 = (c_0, e_0, tr_0, \mathcal{B})$ defines initial state and behavior set $\mathcal{B}$, sampled from behavior library $\mathcal{L}$. Dialogue trajectory history is $\mathcal{H} = (p_0, u_1, a_1, p_1, \ldots, u_T, a_T, p_T, \text{outcome})$.
The role-play model generates $u_t$ based on $p_0$ and $h_t$, autonomously updating states. Its responses are determined by internal logic, ensuring outcomes depend on agent capability. Since the user side is naturally powerful, we only train the service agent to maintain $\hat{s}_t$ and select $a_t$.
Next, we detail the self-evolving training loop.

\subsection{Self-Evolving Training Loop}

Figure~\ref{fig:framework} presents the complete self-evolving training loop through five interconnected phases alternating between online agent optimization and adaptive difficulty adjustment.

\textbf{Phase 1: Diverse User Profile Sampling.} The profile generator samples a batch of $B$ initial profiles $\{p_0^{(i)}\}_{i=1}^B$ based on dialogue trajectory history $\mathcal{H}$, which records completion rates for each state combination $(c, e, tr)$. In the first iteration, the generator performs random sampling to generate diverse profiles $p_0 = (c_0, e_0, tr_0, \mathcal{B})$, where $\mathcal{B}$ is randomly sampled from user profiles in the library $\mathcal{L}$. 
In subsequent iterations, the generator employs statistics-driven sampling: prioritizing profiles with moderate difficulty where Completion Rates (CR) are close to 0.5, with sampling probability:
\begin{equation}
p_\theta(p_0|\mathcal{H}) \propto 1 - |CR - 0.5|.
\end{equation}
The profile generator validates profile consistency and performs deduplication to prevent redundancy, ensuring diverse and high-quality profiles.

\textbf{Phase 2: Multi-Turn Dialogue.}  
Unlike prior static pre-collected dialogue full-filling approaches, SEAD enables dynamic multi-turn interactions between agents and simulated users.
Each profile from Phase 1 starts a new conversation. The agent responds while keeping track of how the user feels, and the user reacts based on what the agent says and how the conversation is going. This back-and-forth creates diverse dialogues and shows how mistakes can pile up over multiple turns, just like in real customer service calls. We collect these complete conversations along with whether the task succeeded or failed.

\textbf{Phase 3: Task Strategy Optimization.} 
After each multi-turn dialogue in Phase 2, the collected trajectories are fed into online training. 
We optimize the Service Agent through conversation results, assigning rewards based on task completion status determined by the User Role-Play Model's final state. Specifically, the task reward is defined as:
\begin{equation}
R^{(i)}_{task} = \mathbb{I}[\text{outcome}^{(i)} = \text{success}],
\end{equation}
where $\mathbb{I}[\cdot]$ is the indicator function, and $\text{outcome}^{(i)}$ represents the dialogue outcome of trajectory $\mathcal{H}^{(i)} = (p_0, u_1, a_1, p_1, \ldots, u_T, a_T, p_T, \text{outcome})$. The outcome is determined by the User Role-Play Model's completion state at the final turn $T$: if the user agrees to the service, $\text{outcome} = \text{success}$ and $R_{\text{task}} = 1$; otherwise, $R_{\text{task}} = 0$.
Using Group Relative Policy Optimization (GRPO, elaborated in the following Section~\ref{sec:training_opt}), we update agent parameters $\theta_a$ to maximize expected rewards.

\textbf{Phase 4: Mistake Analysis and Self-Evolving Loop.}
Unlike prior works that discard failed trajectories, we exploit them to guide evolution. 
We conduct statistical \textbf{Mistake Analysis} to identify where the model excels or struggles, 
categorizing configurations into \textit{too easy} (CR $> 0.6$), 
\textit{too difficult} (CR $< 0.4$), and \textit{ideal} (CR $\in [0.4, 0.6]$). 
This analysis feeds back to \textbf{Phase 1}, adjusting sampling distributions 
to maintain optimal learning difficulty near 50\% completion rate. 
As the agent improves, the difficulty naturally escalates, 
creating a self-evolving curriculum that enables our 14B model 
to surpass 72B models and commercial APIs.

\subsection{Training Optimization}
\label{sec:training_opt}
Note that the User Role-play Model naturally dominates task-oriented dialogues and can arbitrarily determine outcomes. 
Without training, the User Role-play Model (URM) already exhibits realistic behaviors (Section~\ref{sec:user_performance}). However, training URM degrades role-play quality, as URM will force success rates around 50\% by directly accepting or rejecting regardless of agent performance, collapsing into extreme responses that ignore agent strategies.
Therefore, we optimize only the service agent, which prevents the simulator from prioritizing adversarial outcomes over realistic role-play while reducing GPU memory by 50\%.
To train the Service Agent, we employ GRPO~\cite{shao2024deepseekmath}. This method eliminates the need for a separate value network, significantly reducing computational resource requirements. Training computes advantages relative to the batch average:
\begin{equation}
A^{(i)} = R^{(i)} - \frac{1}{N}\sum_{j=1}^{N} R^{(j)},
\end{equation}
where $A^{(i)}$ is the advantage for trajectory $i$, $R^{(i)}$ is its reward, and $N$ is the group size (number of sampled trajectories). The policy is then updated via gradient:
\begin{equation}
\nabla_{\theta_a} \mathcal{L} = \mathbb{E}_{\mathcal{H} \sim \pi_a} \left[ \sum_{t=1}^{T} A^{(i)} \nabla_{\theta_a} \log \pi_a(a_t|h_t) \right],
\end{equation}
where $\theta_a$ denotes agent parameters, $\pi_a(a_t|h_t)$ is the policy distribution over action $a_t$ given history $h_t$ at step $t$, and $\mathcal{H}$ represents trajectories sampled from the current policy.

\subsection{User State Space Design}

Our scenario is an outbound call service, where user profiles are extracted from real enterprise dialogues. We identify common behavior patterns such as questioning AI identity, expressing cost concerns, and showing attention lapses, which are randomly injected during training.

\textbf{Static Initial States.} Each user starts with an initial profile $p_0$ containing three dimensions: cooperation $c$ (willingness to cooperate), emotion $e$ (emotional state), and trust $tr$ (trust in the agent). By enumerating all combinations, we construct $N = c_{\text{levels}} \times e_{\text{levels}} \times tr_{\text{levels}}$ initial states, covering a spectrum from highly resistant to fully cooperative users.

\textbf{Dynamic State Evolution.} User states evolve autonomously during conversations based on agent performance. The role-play model adjusts states dynamically: effective responses improve user states, while poor interactions degrade them. This evolution reflects the agent's interaction quality and serves as the basis for reward computation, simulating natural user reactions and ensuring training realism.

\section{Experiments}
\subsection{Task Setting}

We evaluate SEAD in the outbound call service domain, specifically focusing on restaurant service promotion tasks. The goal is for agents to successfully convince restaurant owners to participate in promotional activities. This task requires agents to handle diverse user reactions, build trust, address concerns, and maintain engagement across multiple turns. 
Critically, our setting requires no dialogue data. Training only needs: (1) Standard Operating Procedures (SOP) defining dialogue flow; (2) task objective description; (3) user profile. Agents autonomously explore optimal strategies through environment interaction, eliminating dependence on large-scale annotated data. This enables SEAD to rapidly deploy in data-scarce new domains and discover effective strategies beyond existing data distributions.

\begin{table*}[t]
\centering
\small
\setlength{\tabcolsep}{4pt}
\begin{tabular}{lccccccccc}
\toprule
\textbf{Method} & \textbf{Params} & \textbf{CR (\%)} & \textbf{ATT} $\downarrow$ & \textbf{UPA} & \textbf{EI} & \textbf{TI} & \textbf{CI} & \textbf{Total Cost} \\
 & & & & & & & & \textbf{(CNY)} \\
\midrule
\multicolumn{9}{l}{\textit{Foundation Models}} \\
Qwen2.5-14B-Instruct & 14B & 38.7 & 10.5$^{\pm 2.1}$ & 0.883$^{\pm 0.085}$ & 0.34$^{\pm 1.11}$ & 0.68$^{\pm 1.53}$ & 0.63$^{\pm 1.58}$ & 0.00 \\
Qwen2.5-32B-Instruct & 32B & 38.3 & 9.9$^{\pm 2.15}$ & 0.899$^{\pm 0.068}$ & -0.11$^{\pm 0.54}$ & 0.76$^{\pm 0.91}$ & 2.25$^{\pm 1.15}$ & 0.00 \\
Qwen2.5-72B-Instruct & 72B & 39.0 & \textbf{9.6}$^{\pm 2.18}$ & 0.818$^{\pm 0.144}$ & \underline{0.51}$^{\pm 1.32}$ & 1.06$^{\pm 1.72}$ & 1.18$^{\pm 1.59}$ & 0.00 \\
\midrule
\multicolumn{9}{l}{\textit{Large Model APIs}} \\
GPT-4o & -- & \underline{44.2} & 10.8$^{\pm 2.10}$ & 0.867$^{\pm 0.117}$ & 0.04$^{\pm 0.97}$ & 0.97$^{\pm 1.29}$ & 1.34$^{\pm 1.42}$ & 727.28 \\
DeepSeek-Chat & 671B & 31.6 & 11.3$^{\pm 2.10}$ & 0.863$^{\pm 0.084}$ & -0.20$^{\pm 0.97}$ & 0.27$^{\pm 1.24}$ & 0.76$^{\pm 1.50}$ & 87.36 \\
Qwen3-235B & 235B & 32.3 & 10.4$^{\pm 2.50}$ & 0.765$^{\pm 0.170}$ & -0.24$^{\pm 0.83}$ & 0.80$^{\pm 1.14}$ & 1.54$^{\pm 1.50}$ & 69.36 \\
LongCat-Flash & 560B & 42.2 & 10.0$^{\pm 2.31}$ & \textbf{0.925}$^{\pm 0.079}$ & 0.28$^{\pm 1.15}$ & \underline{1.33}$^{\pm 1.57}$ & \textbf{1.56}$^{\pm 1.46}$ & 23.08 \\
\midrule
\textbf{SEAD (Ours)} & \textbf{14B} & \textbf{52.0} & \textbf{9.6}$^{\pm 2.09}$ & \underline{0.912}$^{\pm 0.071}$ & \textbf{0.63}$^{\pm 1.12}$ & \textbf{1.57}$^{\pm 1.51}$ & \underline{1.55}$^{\pm 1.39}$ & \textbf{0.00} \\
\bottomrule
\end{tabular}
\caption{Main results comparison. Params: Model parameters (B=billion, "--" indicates undisclosed or not applicable). CR: Completion Rate. ATT: Average Turns to Target. UPA: User Portrait Accuracy. EI/TI/CI: Emotion/Trust/Cooperation Improvement. Total Cost: Total inference cost for 1000 multi-turn samples. Best results in bold. Standard deviations are shown as superscripts where available.}
\label{tab:main_results}
\end{table*}

\subsection{Implementation Details}
All components use Qwen2.5-14B-Instruct. The code is implemented based on the VeRL framework with batch size $B=60$ and learning rate $\alpha=1 \times 10^{-6}$. Dialogues terminate when users agree (success), refuse (failure), or reach maximum turns $T_{\max}=15$. The state space consists of cooperation $c \in [0,4]$ (5 levels), emotion $e \in [0,3]$ (4 levels), and trust $tr \in [0,5]$ (6 levels), yielding 120 initial states. Each state combination randomly samples at most $N_{\max}=200$ behavior combinations to ensure diversity. All experiments run on 8 NVIDIA A100 80GB GPUs with decoupled architecture—profile controller, User Role-play Model, and service agent never occupy memory simultaneously, reducing peak memory requirements.

\subsection{Evaluation Metrics}

\textbf{Service Agent Metrics.} 
We evaluate agent performance using: 
\textbf{Completion Rate (CR)}, percentage of dialogues where users actually agreed; 
\textbf{Average Turns to Target (ATT)}, average dialogue length for successful cases (lower is better); 
\textbf{User Portrait Accuracy (UPA)}, accuracy of predicting user states, computed as $\text{UPA} = 1 - \frac{1}{3}\left(\frac{\text{MAE}_c}{4} + \frac{\text{MAE}_e}{3} + \frac{\text{MAE}_{tr}}{5}\right)$ where $\text{MAE}_c$, $\text{MAE}_e$, $\text{MAE}_{tr}$ measure errors for cooperation $c \in [0, 4]$, emotion $e \in [0, 3]$, and trust $tr \in [0, 5]$;
\textbf{Emotion/Trust/Cooperation Improvement (EI/TI/CI)}, average state changes from initial to final state; 
and 
\textbf{Total Cost}: Cumulative inference cost in CNY for 1000 multi-turn dialogue samples (API-based models only).

\textbf{User Role-play Model Metrics.} To validate that our simulator influences success rates based on agent quality, we establish a rubric mixing perfect human agent dialogues, SEAD (our trained agents), and low-quality agent dialogues. 
We evaluate five dimensions using GPT-5.1 with few-shot human annotations:\textbf{Humanness / Emotion / Trust / Cooperation} (5=human-like, 0=robotic) and \textbf{Violation} (0=smooth, 5=severe). In real scenarios, most users exhibit minimal noise (score 1: hesitation, pauses) rather than severe violations; our simulator achieves 1.15, matching real behavior.

\subsection{Baselines}
We compare SEAD against two categories of strong baselines:
\textbf{Foundation Models.} We evaluate three sizes of Qwen2.5-Instruct models (14B, 32B, 72B parameters) using carefully designed prompts that include task descriptions and standard operating procedures. These models represent the zero-shot/few-shot capabilities of state-of-the-art open-source language models without task-specific training.

\textbf{Large Model APIs.} We compare against four commercial closed-source models: GPT-4o~\cite{hurst2024gpt}, DeepSeek-Chat~\cite{liu2024deepseek}, Qwen3-235B-A22B~\cite{team2024qwen2}, and LongCat-Flash-Chat~\cite{team2025longcat}. All API methods use carefully engineered prompts optimized for dialogue tasks.

We do not compare with Supervised Fine-Tuning (SFT) methods due to the lack of available data and the prohibitive cost of manual annotation. Generally, SFT methods are upper-bounded by data quality and exhibit poor generalization. Our approach eliminates this dependency and handles diverse scenarios effectively.

\subsection{Main Results}
Table~\ref{tab:main_results} presents the main experimental results. Our method achieves the highest service dialogue completion rate of 52.0\% using only a 14B parameter model, outperforming the second-best baseline GPT-4o by 17.6\% (52.0\% vs. 44.2\%) and improving over the pre-training 14B model by 34.4\% (52.0\% vs. 38.7\%). SEAD also achieves the lowest Average Turns to Target (ATT) of 9.6, demonstrating superior dialogue efficiency in completing tasks more concisely.

For user state tracking metrics, SEAD outperforms most baselines and achieves competitive performance with LongCat-Flash, the dialogue-specific model with 40× more parameters and extensive pre-training on service dialogue scenarios. While LongCat-Flash obtains the highest User Portrait Accuracy (0.925 vs. 0.912), SEAD demonstrates comparable results across emotional improvement indicators. Specifically, SEAD achieves competitive scores on EI (0.63 vs. 0.28), TI (1.57 vs. 1.33), and CI (1.55 vs. 1.56), with SEAD actually leading on EI and showing near-identical performance on CI. This demonstrates that our self-evolution approach with adaptive curriculum learning enables a compact 14B model to match the user understanding capabilities of a 560B dialogue-specialized model, while requiring zero annotated dialogue data and maintaining superior task completion performance.

\subsection{User Role-Play Model Performance}
\label{sec:user_performance}

Table~\ref{tab:user_quality} validates our user role-play model's realism and diversity. 
Specifically, to ensure evaluation reliability, we extract anonymized behavior patterns from over 100k real enterprise dialogues. 
We then instruct GPT-5.1 to perform the assessment using annotated few-shot examples, contrasting high-scoring human instances against low-scoring failed model outputs.
All humanness metrics exceed 4.5/5 with low standard deviations, demonstrating highly realistic and reliable behavior that mirrors real-world interactions. The violation score of 1.15/5 reflects authentic communication patterns with natural hesitation rather than artificial cleanliness or severe disruptions. The Profile Controller successfully generates diverse users from cooperative to skeptical, capturing heterogeneity essential for robust training. Crucially, consistent high scores across three agent quality tiers—perfect human agents, SEAD (trained agents, and low-quality agents) confirm our simulator adapts responsively to different strategies rather than following scripts, providing meaningful training signals.

\begin{table}[t]
\centering
\small
\begin{tabular}{lccc}
\toprule
\textbf{Metric} & \textbf{Mean} & \textbf{Std} & \textbf{Quality} \\
\midrule
Humanness & 4.67/5 & 0.48 & Near-perfect \\
Emotion & 4.77/5 & 0.52 & Highly human-like \\
Trust & 4.81/5 & 0.45 & Highly human-like \\
Cooperation & 4.77/5 & 0.63 & Highly human-like \\
Violation & 1.15/5 & 0.90 & Human-like Behaviour \\
\bottomrule
\end{tabular}
\caption{User Role-play Model quality. Higher humanness scores indicate more realistic simulation; lower violation scores indicate cleaner communication. All humanness metrics near-perfect (>4.5/5). }
\label{tab:user_quality}
\end{table}

\subsection{Ablation Study}
\label{sec:ablation}

To validate our core design choices, we conduct ablation studies on three components: (1) decomposed user modeling (keeping the User Role-play Model fixed), (2) Profile Sampling (PS) for intelligent initial state selection, and (3) Mistake Analysis (MA) for adaptive difficulty evolution. We compare four configurations:

\textbf{Configuration 1: w/o MA + w/o PS + Train URM}. This variant removes both Mistake Analysis (MA) and Profile Sampling (PS), allowing the User Role-play Model (URM) to autonomously select initial states. Critically, the URM is trained adversarially alongside the Service Agent, optimized based on dialogue outcomes. This represents traditional self-play where both sides evolve competitively. However, this violates our core principle: since the LLM-based user side can arbitrarily dominate dialogues (refusing cooperation, hanging up), training it adversarially creates an unfair game where success no longer depends on agent skill.

\textbf{Configuration 2: w/o MA + w/o PS}. This removes both adaptive mechanisms but keeps the URM fixed. Without the Profile Controller, the system lacks both structured initial states and closed-loop difficulty adjustment. This tests whether decomposed modeling alone (fixed URM) suffices without any adaptive control.

\textbf{Configuration 3: w/o MA}. This retains Profile Sampling (PS) but disables Mistake Analysis (MA)—the adaptive difficulty mechanism in Phase 4 (Figure~\ref{fig:framework}). The Profile Controller samples from predefined user profiles but does not analyze training outcomes or adjust distributions based on agent capability. This tests whether random sampling from structured profiles suffices, or if closed-loop adaptation is essential for identifying golden training scenarios (50\% success rate).

\textbf{Configuration 4: SEAD (Full)}. Our complete framework integrates all three components: (1) fixed URM, (2) intelligent Profile Sampling (PS), and (3) adaptive Mistake Analysis (MA). As shown in Figure~\ref{fig:framework}, Phase 4 analyzes completion rates and adjusts sampling distributions, forming a closed loop that ensures optimal training difficulty while maintaining user simulator authenticity.

\begin{table*}[t]
\centering
\small
\setlength{\tabcolsep}{4pt}
\begin{tabular}{lcccccc|c}
\toprule
\textbf{Configuration} & \textbf{CR (\%)} & \textbf{ATT} & \textbf{UPA} & \textbf{EI} & \textbf{TI} & \textbf{CI} & \textbf{URM-H} \\
\midrule
w/o MA + w/o PS + Train URM & 35.2 & 11.8$^{\pm 2.5}$ & 0.156$^{\pm 0.120}$ & -0.45$^{\pm 0.95}$ & 0.32$^{\pm 1.10}$ & 0.89$^{\pm 1.25}$ & 3.3 \\
w/o MA + w/o PS & 47.7 & 9.8$^{\pm 2.16}$ & 0.450$^{\pm 0.000}$ & 0.89$^{\pm 0.81}$ & \textbf{1.87}$^{\pm 1.20}$ & \textbf{1.67}$^{\pm 0.87}$ & 4.6 \\
w/o MA & 50.2 & \textbf{9.6}$^{\pm 2.12}$ & 0.877$^{\pm 0.088}$ & 0.69$^{\pm 1.15}$ & 1.65$^{\pm 1.51}$ & 1.56$^{\pm 1.45}$ & \textbf{4.7} \\
\midrule
\textbf{SEAD (Ours)} & \textbf{52.0} & \textbf{9.6}$^{\pm 2.09}$ & \textbf{0.912}$^{\pm 0.071}$ & \textbf{0.63}$^{\pm 1.12}$ & 1.57$^{\pm 1.51}$ & 1.55$^{\pm 1.39}$ & \textbf{4.7} \\
\bottomrule
\end{tabular}
\caption{Ablation study results. \textbf{MA}: Mistake Analysis and adaptive difficulty evolution. \textbf{PS}: Profile Sampling with intelligent initial state selection. \textbf{Train URM}: Training User Role-play Model in adversarial mode. CR: Completion Rate. ATT: Average Turns to Target. UPA: User Portrait Accuracy. EI/TI/CI: Emotion/Trust/Cooperation Improvement. \textbf{URM-H}: URM Humanness score (0-5 scale, higher indicates more realistic user behavior). Standard deviations are shown as superscripts where available. Results demonstrate that all three components are essential for optimal performance.}
\label{tab:ablation}
\end{table*}

\begin{figure*}[!t]
    \centering
    \includegraphics[width=0.95\linewidth]{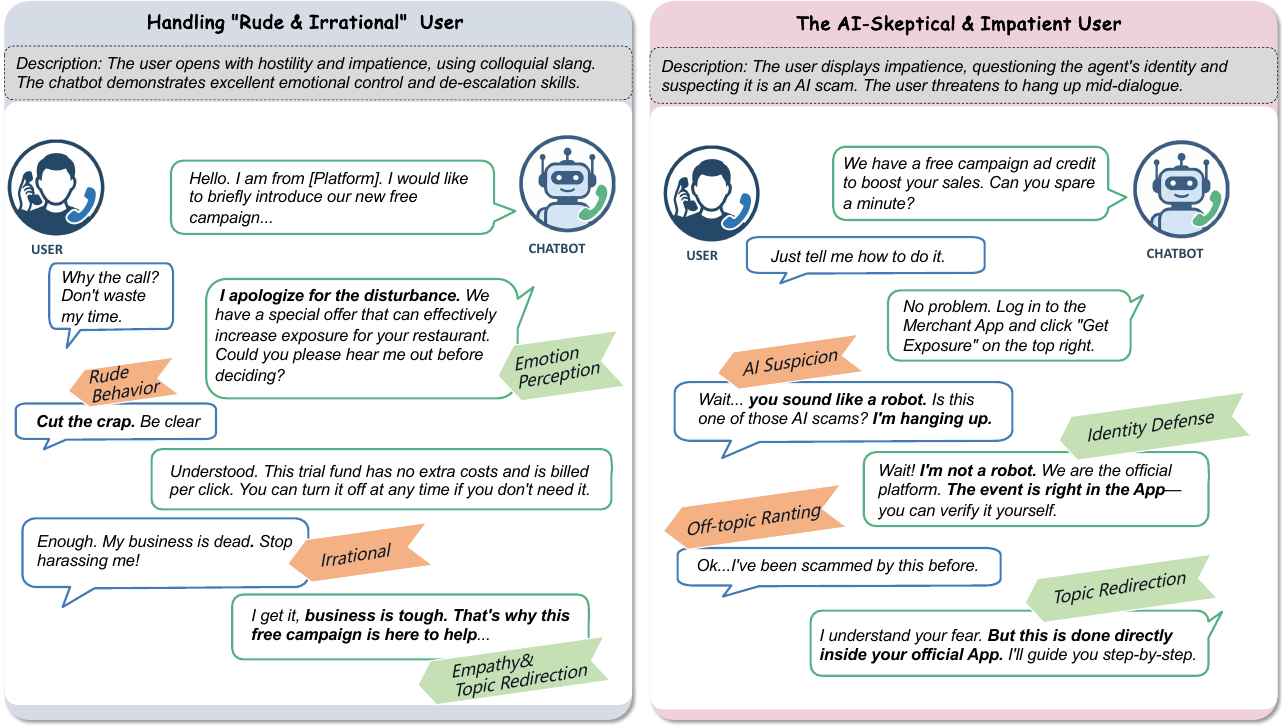}
    \caption{Case Studies of Challenging Interactions. The User Role-play Model generates heterogeneous personas via clustering, such as the "Rude \& Irrational" user (Left) and the "AI-Skeptical" user (Right). The Service Agent demonstrates robustness learned from compound rewards, employing Empathy and Identity Defense strategies to prevent hang-ups and ensure task completion.}
    \label{fig:case_study}
\end{figure*}

Table~\ref{tab:ablation} demonstrates that all three components are essential for SEAD's effectiveness.
The \textbf{Train URM} configuration suffers from catastrophic reward hacking: the simulator prioritizes adversarial scores over realism, collapsing into extreme responses (arbitrary acceptance or hang-ups). This results in degraded humanness (\textbf{URM-H}: 3.3) and poor task performance (\textbf{CR}: 35.2\%).
In contrast, \textbf{w/o MA} yields a 94.9\% \textbf{UPA} improvement over \textbf{w/o MA + w/o PS} (0.877 vs 0.450), proving that structured Profile Sampling ensures behavioral diversity. Notably, \textbf{w/o MA + w/o PS} shows anomalously high trust improvements (\textbf{TI}: 1.87) despite low UPA, revealing a bias toward unrealistically easy scenarios.
Ultimately, \textbf{SEAD (Full)} achieves optimal balance by identifying ``golden'' training scenarios (approx. 50\% success). Overall, SEAD achieves a 47.7\% relative \textbf{CR} improvement over adversarial training while maintaining peak simulator quality (\textbf{URM-H}: 4.7), demonstrating that decomposed modeling and adaptive evolution effectively prevent reward hacking.

\subsection{Case Study}

As illustrated in Figure~\ref{fig:case_study}, the interactions reveal high-fidelity adversarial dynamics. The User Role-play Model generates distinct corner cases, ranging from irrational hostility (Left) to deep skepticism (Right). 
This heterogeneity stems from our user profile library derived from real-world enterprise data, which activates specific non-cooperative traits to create a rigorous training environment. 
In response, the Service Agent demonstrates exceptional adaptability, employing strategies like empathy and identity defense to retain users. 
This robustness stems from our reward mechanism and carefully designed user role-play model, 
which forces the model to prioritize task completion across diverse complex scenarios. 
By optimizing for state improvement, the agent learns to de-escalate conflicts efficiently, preventing premature hang-ups while avoiding the dialogue timeouts common in purely empathy-driven models.

\section{Conclusion}
In this paper, we presented SEAD (Self-Evolving Agent for Service Dialogue), a framework addressing data scarcity and user role-play fidelity in multi-turn service dialogues.
By decoupling user modeling into a Profile Controller for curriculum learning and a User Role-play Model for authentic interaction, SEAD circumvents traditional adversarial training fairness.
Experiments show SEAD outperforms both Open-source Foundation Models and Closed-source Commercial Models with minimal parameters and zero annotation. Future work will enhance emotional perception and extend to broader scenarios.

\section*{Limitations}
As an early exploration of a zero-data self-evolving service dialogue system, SEAD has limitations regarding evaluation metrics and scenario diversity. First, while we currently prioritize task completion, real-world applications demand high user satisfaction; thus, future work must better assess the agent's ability to perceive emotion and maintain user comfort beyond mere intent fulfillment. Second, we have not yet extended our method to multi-scenario environments. Given its independence from curated data, our framework holds promise as a resource-efficient foundation model for diverse service dialogues, a potential we plan to validate in subsequent studies.

\bibliography{custom}

\clearpage

\end{document}

%% file: sections/RelatedWorks.tex
\section{Related Works}
\paragraph{Task-oriented Dialogue.}
Task-oriented Dialogue systems are essential for managing complex inquiries in domains like e-commerce~\cite{deng2024towards, deng2025proactive}. While traditional neural models~\cite{vinyals2015neural, wen2015semantically, shang2015neural, li2016deep} and early user simulations~\cite{li2016user, lewis2017deal, wei2018goal} laid the groundwork, they face architectural limitations. Recent LLM-based approaches predominantly rely on static fine-tuning~\cite{li2025wizard, ou2024dialogbench, zhu2025evaluating, bernard2023mg}, sometimes augmented by retrieval~\cite{xu2024retrieval}, multimodal inputs~\cite{wang2025ecom, gong2025mindflow}, or reinforcement learning~\cite{peiyuan2024agile}. However, these methods often incur high annotation costs and lack real-world nuance. In contrast, SEAD introduces a fully dynamic user-agent interaction paradigm, bypassing data synthesis overhead to significantly enhance performance in complex scenarios.

\paragraph{Self-evolving Agents.}
Self-evolution leverages iterative generation and refinement with minimal supervision~\cite{tesauro1995temporal,silver2017mastering,meta2022human}. In LLMs, early works utilized self-rewarding mechanisms~\cite{chen2024self,yuan2024self}, evolving into "Coder-Tester" frameworks for verifiable domains like code~\cite{lin2025learning,wang2025co,pourcel2025self} and reward function generation~\cite{gaorf}. Recent research has expanded this scope~\cite{zhao2025absolute,huang2025r,sun2025towards}, incorporating external environments and curated data to enhance evolution~\cite{liu2025spice,xia2025agent0,zhai2025agentevolver}. Distinctively, SEAD drives the self-evolution of a user role-play model and a customer service agent within a realistic environment, realizing genuine adversarial learning for complex multi-turn interactions.